\documentclass[runningheads]{llncs}
\usepackage[T1]{fontenc}
\usepackage{graphicx}
\usepackage[pagebackref,breaklinks,colorlinks]{hyperref}

\usepackage{amsmath}
\usepackage[capitalize]{cleveref}
\crefname{section}{Sec.}{Secs.}
\Crefname{section}{Section}{Sections}
\Crefname{table}{Table}{Tables}
\crefname{table}{Tab.}{Tabs.}

\newcommand{\etal}{\textit{et al}. }

\newcommand{\ie}{\textit{i}.\textit{e}.}
\newcommand{\eg}{\textit{e}.\textit{g}.}

\usepackage{enumitem}

\newcommand{\RN}[1]{\uppercase\expandafter{\romannumeral#1}}
\newcommand{\Rn}[1]{\romannumeral#1\relax}

\usepackage{stix}

\usepackage{booktabs}
\usepackage{adjustbox}
\usepackage{array}
\usepackage{longtable,lscape}

\usepackage{multirow}
\usepackage{hhline}
\usepackage{listings}

\usepackage{pifont}
\begin{document}
\title{From Local Binary Patterns to Pixel
Difference Networks for Efficient Visual Representation Learning}
\titlerunning{From LBPs to Pixel Difference Networks for Effi. Vis. Rep. Learning.}
%

\author{Zhuo Su\inst{1} 
\and
Matti Pietik{\"a}inen\inst{1} \and
Li Liu\inst{2,1,}\thanks{Corresponding author: li.liu@oulu.fi}
}
\authorrunning{Z. Su, L. Fang et al.}
%
\institute{Center for Machine Vision and Signal Analysis, University of Oulu, Finland \and
National University of Defense Technology, China
}
%
\maketitle              
\begin{abstract}
LBP is a successful hand-crafted feature descriptor in computer vision. However, in the deep learning era, deep neural networks, especially convolutional neural networks (CNNs) can automatically learn powerful task-aware features that are more discriminative and of higher representational capacity. To some extent, such hand-crafted features can be safely ignored when designing deep computer vision models. Nevertheless, due to LBP's preferable properties in visual representation learning, an interesting topic has arisen to explore the value of LBP in enhancing modern deep models in terms of efficiency, memory consumption, and predictive performance. In this paper, we provide a comprehensive review on such efforts which aims to incorporate the LBP mechanism into the design of CNN modules to make deep models stronger. In retrospect of what has been achieved so far, the paper discusses open challenges and directions for future research.

\end{abstract}
\section{Introduction}
Local binary pattern (LBP) is one of the most prominent feature descriptors in the computer vision community. With its distinctive advantages, namely, ease of implementation, invariance to monotonic illumination changes, and low computational complexity, it was widely studied with numerous LBP variants proposed and applied for a diverse range of applications such as texture classification~\cite{ojala1996lbp1996,pietikainen2000lbp2000ri,ojala2002lbp2002riu2}, dynamic texture recognition~\cite{zhao2007lbp-top}, image matching~\cite{heikkila2009app-imagematching}, visual inspection~\cite{silven2003app-visualinspection}, image retrieval~\cite{doshi2012app-imageretrevial}, biomedical image analysis~\cite{nanni2010app-biomedical1,nanni2010app-biomedical2}, facial analysis~\cite{ahonen2004facelbp,zhang2005localgabor,tan2010ltp}, motion and activity analysis~\cite{kellokumpu2008app-motion}, object detection~\cite{trefny2010app-objectdetection1,satpathy2014app-objectdetection2}, and background subtraction~\cite{heikkila2006app-background}. 

With the development of deep learning~\cite{DBLP:conf/NeurIPS/12alexnet}, convolutional neural networks (CNNs), on the other hand, appeared to be a more powerful feature extractor for visual inputs. Instead of hand-crafting features like LBP, CNN structures are capable of learning rich hierarchical features from local texture statistics to global semantic information that helps the models to achieve human-level or even surpass-human performances in many visual applications. In the deep learning era, it is necessary to think about the following questions: \emph{Is LBP still worth exploring in computer vision?} \emph{What role does LBP take when CNNs are so strong?} or \emph{Can LBP help the design of CNNs to make our models stronger and how?}

\begin{figure}[t!]
\centering
    \centering
    \includegraphics[width=\linewidth]{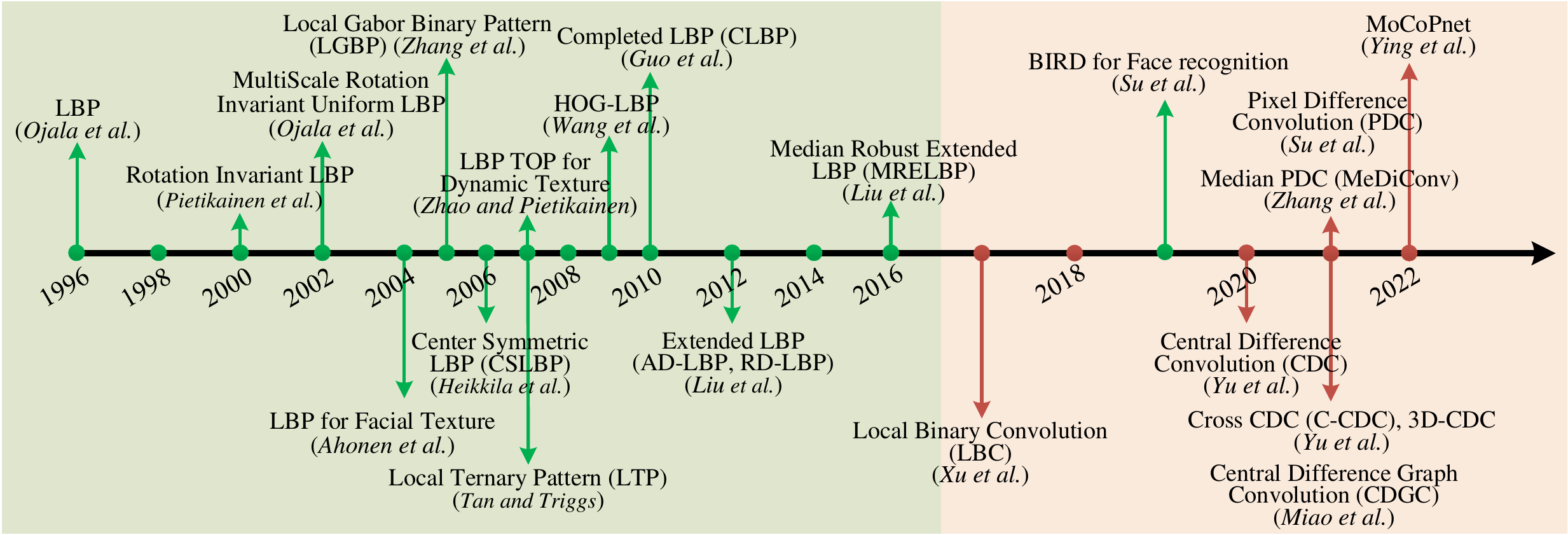}
    \caption{The evolution of LBP over the past decades.}
    \label{fig:history}
\end{figure}

In this paper, we aim to answer these questions with a review of the recent efforts on LBP in the deep learning era. Briefly, LBP, or the design philosophy of LBP, has still shown great benefits in the past years that help CNN models to boost their performance in terms of prediction accuracy, efficiency, and memory, on a wide range of applications like face anti-spoofing, edge detection, gesture recognition, and object classification. We hope our review can inspire researchers seeking a higher-level perspective in the future. A brief history of LBP is depicted in~\cref{fig:history}.

The rest of the paper is organized as follows. In~\cref{sec:lbp}, we give a short introduction of traditional LBP and its variants. Then, we focus our attention on the works of LBP-inspired CNN modules in~\cref{sec:cnn-lbp}. Applications of the introduced methods are present in~\cref{sec:app}. Finally, we discuss possible future works and conclude the paper in~\cref{sec:conclusion}.

\section{Traditional LBP}

\begin{figure}[t!]
\centering
    \centering
    \includegraphics[width=\linewidth]{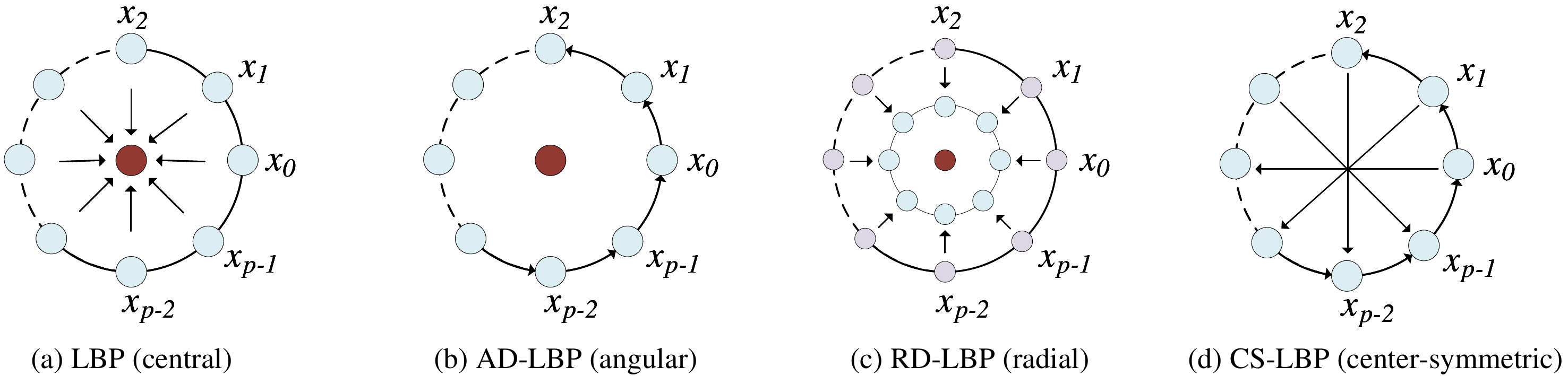}
    \caption{Illustration of the original LBP~\cite{ojala1996lbp1996}, angular LBP~\cite{liu2012extended}, radial LBP~\cite{liu2012extended}, and center-symmetric LBP~\cite{heikkila2006cslbp}.}
    \label{fig:lbp}
\end{figure}

\label{sec:lbp}
LBP descriptors were firstly introduced by Ojala \emph{et.al.}~\cite{ojala1996lbp1996,ojala2002lbp2002riu2} to encode pixel-wise information in textured images. 
Specifically, an input image is probed locally by sampling the values from the neighborhood. As shown in Fig.~\ref{fig:lbp} (a), for a certain pixel $x_c$, the values from neighboring locations $\{x_0, x_1, ..., x_{p-1}\}$ spaced equidistantly around a circle with radius $r$ are extracted to generate a binary code composed of 0 and 1, by comparing each of those values with the central value $x_c$. That is, the neighboring values greater than or equal to $x_c$ are associated with 1, otherwise with 0. The binary values 0 and 1 are read anticlockwise from the starting point $x_0$ to the ending point $x_{p-1}$, leading to a $p$-length binary code as a descriptive local pattern, which is then represented by a decimal number. The feature of the image is then represented by the histogram of all the possible binary codes. The generation of an LBP code can be formulated as:

\begin{equation}
    \text{LBP}_{r,p}(c) = \sum_{i=0}^{p-1}s(x_i - x_c)2^i, \;\;\; s(x) = \begin{cases}
        1\;\;  x\geq 0,\\
        0\;\;  x < 0.
    \end{cases}
    \label{eq:lbp}
\end{equation}

By altering the number of sampling pixels $p$ and the radius $r$, LBP patterns can be extracted in different scales and complexities. 

While simple in computation and invariant to illumination changes, the original LBP (\cref{fig:lbp} (a)) suffers from significant disadvantages such as exponential growth of possible patterns with the increase of $p$ (\ie, $2^p$), failure to detect large-scale texture structures, sensitivity to image rotation and noise. To address the above limitations, numerous LBP variants were successfully developed such as the rotation invariant LBP~\cite{pietikainen2000lbp2000ri,60784312012lbprotationvideo}, uniform LBP~\cite{ojala2002lbp2002riu2}, extended LBP~\cite{liu2012extended} (\cref{fig:lbp} (b-c)), center-symmetric LBP (\cref{fig:lbp} (d)), noise robust LBP~\cite{liu2016mrelbp}. Generally, these methods improve the original LBP from the following aspects: (1) Changing neighborhood topology and sampling~\cite{liu2012extended,heikkila2006cslbp} (\ie, the layout of the sampled neighboring pixels). (2) Adopting different thresholding and quantization methods~\cite{tan2010ltp} (\ie, the implementation of the sign function $s(x)$ in~\cref{eq:lbp}); (3) Designing different encoding or grouping strategies (\eg, rotation invariant patterns~\cite{pietikainen2000lbp2000ri}, uniform patterns~\cite{ojala2002lbp2002riu2}, learnable LBP patterns~\cite{su2019bird}); (4) Combining complementary features (\eg, HOG features~\cite{wang2009hoglbp}, macro-textons with Gabor filters~\cite{ahonen2009filterbanklbp,zhang2005localgabor}, and pixel difference magnitudes~\cite{guo2010clbp}). A more comprehensive review can be seen in~\cite{liu2017lbpsurvey}.

\section{LBP meets CNNs: to benefit from both worlds}
\label{sec:cnn-lbp}

Deep learning has revolutionized computer vision in a broad range of applications. The convolution operators act as the basic local feature descriptors like LBP that probe local details in their current receptive fields. By stacking multiple layers of convolutional layers, the receptive fields of convolution modules are gradually increased, allowing CNN models to effectively capture both fundamental low-level features like textures, colors, corners, and higher-level abstract features that represent object semantics. In contrast, LBP behaves like convolution operators but probes local details in a different way~\cite{ahonen2009filterbanklbp}. 

\setcounter{footnote}{0} 

In the deep learning era, LBP may perform better than CNN models in certain properties.\footnote{It should be noted LBP has possibly more than these three properties. Here we only focus on properties that give more inspirations for designing CNN modules.} In some cases, LBP variants even outperform CNNs for standard (traditional) texture test sets, \eg, in noisy conditions~\cite{liu2016evaluationeccv,liu2017lbpsurvey}. We list the properties as below:
\begin{enumerate}[label=\roman*., font=\itshape]
    \item Computational simplicity and efficiency: LBP codes are binary, thus can be efficiently computed and are memory friendly.
    \item Easing the encoding of higher-order information: the pixel differences within local areas contain rich image gradient cues compared with the original pixel intensities used in CNNs~\cite{su2021pdc}.
    \item Ability of probing various microstructures from images: the numerous neighboring topologies and sampling strategies in LBP and its variants provide diversity to probe rich local patterns for visual inputs.
\end{enumerate}

However, unlike CNNs, LBP descriptors have limited representational capacity due to their fixed ways to calculate the patterns and the associated shallow structures. Since LBP and CNNs are complementary in these properties, how to combine them to benefit from both worlds is an inspiring and meaningful research topic. In the following sections, we discuss some recent efforts under this direction regarding the properties \emph{\Rn{1}.}-\emph{\Rn{3}.} of LBP, and hope to give helpful insights for future works.

\begin{figure}[t!]
\centering
    \centering
    \includegraphics[width=\linewidth]{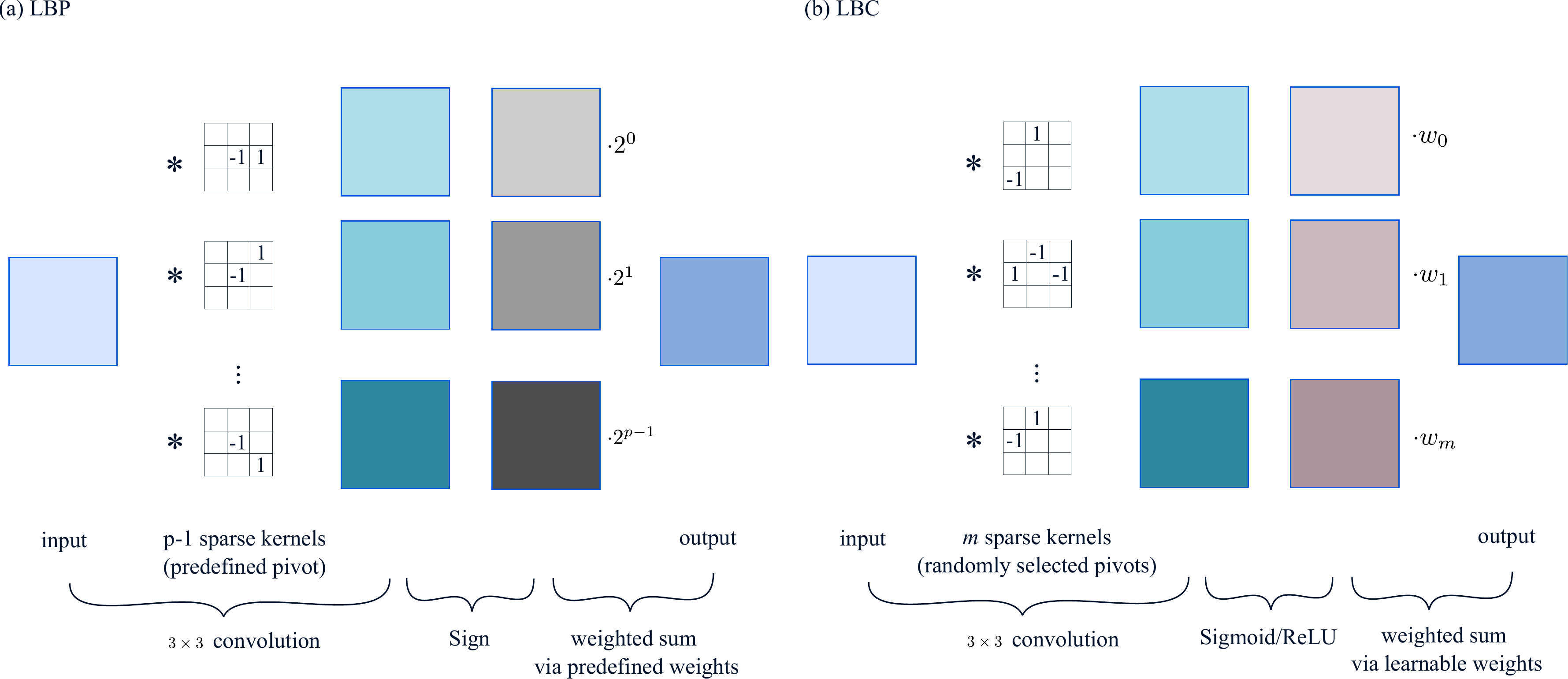}
    \caption{We can regard LBP as a $3\times 3$ convolutional layer, where the calculation of pixel differences between neighboring pixels and the central pixel is equivalent to a series of $3\times 3$ convolutions with sparse binary kernels. After the convolution, the difference maps are process with sign function followed by linear combination via predefined weights ($2^i$). LBC upgrades LBP with alternative sparse binary kernels, Simoid and ReLU functions, and learnable weights to pool those difference maps.}
    \label{fig:lbc}
\end{figure}

\subsection{Local binary convolution}

A pioneering work combing LBP and convolutional operations is local binary convolution (LBC)~\cite{juefei2017lbc,zhang2017rotationlbcnn}. The authors decomposed the pattern generation process of LBP into sub-steps. As implied in~\cref{eq:lbp}, first, the central pixel $x_c$ in a patch is chosen as a pivot to calculate the differences with other neighboring pixels $\{x_i\}_{i=0}^{p-1}$. Second, the nonlinear Sign function is used to convert each pixel difference to a binary value. Finally, all these binary values are pooled via a linear combination with fixed weights $\{2^i\}_{i=0}^{p-1}$. Thereby, LBP can be generalized into a more flexible form by changing the selection of pivot pixel, the nonlinear function, and values of the weights. 

To help embed the above process into the convolutional operation in CNNs, the calculation of differences between pivot pixel and other pixels can be regarded as a $3\times 3$ convolution, where each $3\times 3$ kernel has sparse binary values~\cite{ahonen2009filterbanklbp}, as shown in~\cref{fig:lbc} (a). Based on that, LBC randomly generated $m$ sparse binary kernels with different locations of $-1$ and $1$ (the binary kernels were then fixed afterwards). The locations of $-1$ represent pivot positions. Then, the Sign function was changed to Sigmoid or ReLU to create real-valued maps. Finally, the predefined weights (using base 2) were replaced with learnable real-valued weights to pool these real-valued maps. The process of LBC is illustrated in~\cref{fig:lbc} (b).

\subsubsection{Discussion}
Compared with standard convolution, LBC shows considerable reduction in computational cost and memory storage of the parameters. On one hand, the sparse binary kernels in LBC allow it to generate $m$ intermediate feature maps efficiently, since binary operations are much faster than real-valued counterparts~\cite{courbariaux2016bnn}. On the other hand, linear combination of those intermediate features with learnable weights can be regarded as an efficient $1\times 1$ convolution. Both binary kernels and $1\times 1$ kernels need much less memory storage than standard $3\times 3$ kernels, leading to lower model size.

\begin{figure}[t!]
\centering
    \centering
    \includegraphics[width=\linewidth]{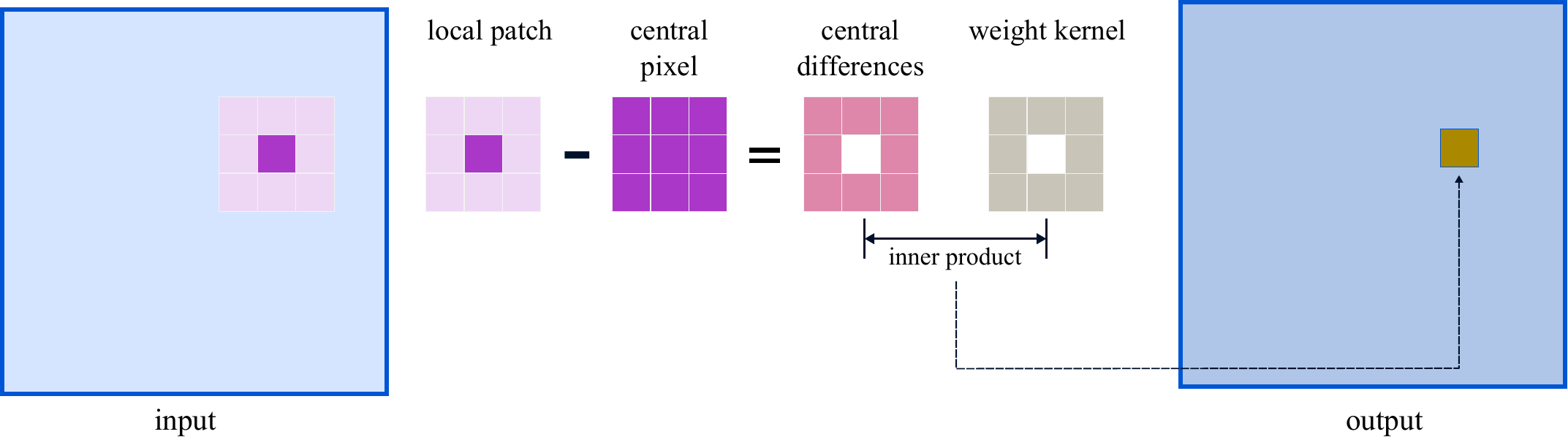}
    \caption{Unlike standard convolution using pixel intensities, CDC leverages central pixel differences to conduct convolutional operation.}
    \label{fig:cdc}
\end{figure}

\subsection{Central difference convolution and its variants}

\subsubsection{Central difference convolution (CDC)}
CDC~\cite{yu2020cdc,yu2020cdc-pami,yu2020cdc-cvprw} incorporates the original LBP into convolutional operation to capture gradient information from the input images. As shown in~\cref{fig:cdc}, to generate a output value in a certain location, a convolution operator firstly samples a local region in the input feature map, consisting of the central pixel $x_c$ and all its neighboring $p$ pixels $\{x_i\}_{i=0}^{p-1}$. Similar to LBP, CDC calculates the differences between the central pixel and its neighboring pixels as central differences, which are aggregated via the learnable kernel weights. Formally, CDC can be written as:

\begin{equation}
    y_c = \sum_{i=0}^{p-1}w_i\cdot (x_i - x_c),
    \label{eq:cdc}
\end{equation}
where $y_c$ is the output value, $\{w_i\}_{i=0}^{p-1}$ represent the kernel weights.

CDC was originally proposed for anti-spoofing task. However, both intensity level semantic information and gradient level text details are crucial for distinguishing live and spoofing faces. Therefore, the authors in~\cite{yu2020cdc} further proposed to combine standard convolution and CDC to create a generalized CDC operator to capture both types of information:

\begin{align}
    y_c = \theta \cdot \underbrace{\sum_{i=0}^{p-1}w_i\cdot (x_i - x_c)}_{\textit{gradient level}} + (1-\theta)\cdot \underbrace{(\sum_{i=0}^{p-1}w_i\cdot x_i + w_c\cdot x_c)}_{\textit{intensity level}},
    \label{eq:gcdc}
\end{align}
where $\theta$ is a hyperparameter that tradeoffs the contribution between intensity level and gradient level information.

\begin{figure}[t!]
\centering
    \centering
    \includegraphics[width=\linewidth]{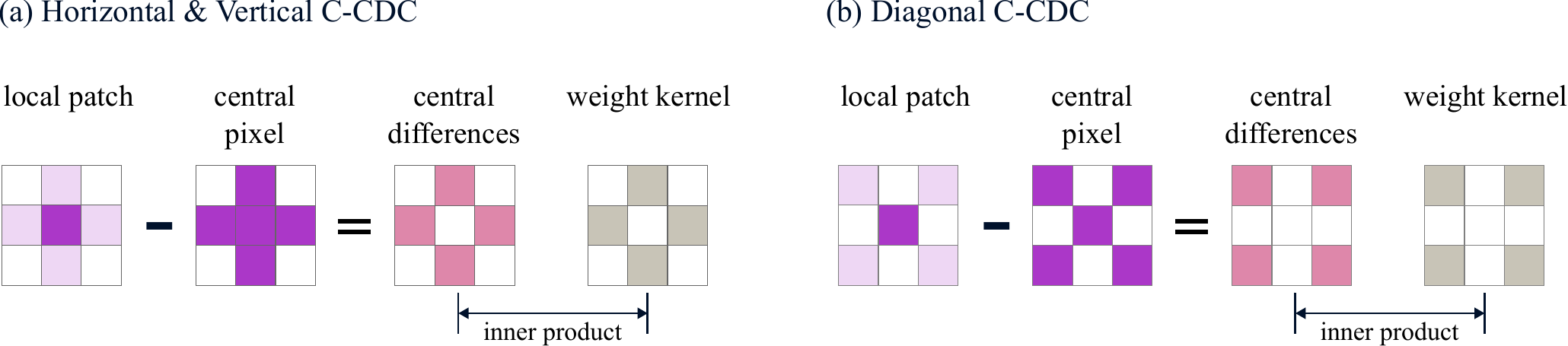}
    \caption{By changing the sampling locations, CDC are revised to two lightweight C-CDC versions.}
    \label{fig:c-cdc}
\end{figure}

\subsubsection{Cross CDC}
In CDC, central gradients are calculated from all neighbors, which might be redundant and sub-optimal due to the discrepancy among diverse gradient directions. Yu \etal~\cite{yu2021c-cdc} proposed to decouple such directions into two cross components (\ie, horizontal/vertical and diagonal) for better extracting gradient information, deriving two types of cross CDC (C-CDC), as shown in~\cref{fig:c-cdc}. Similar to CDC, by combining standard convolution, C-CDC can be generalized to capture both intensity and gradient level information:

\begin{equation}
    y_c = \theta \cdot \underbrace{\sum_{i=0}^{p_{S}-1}w_i\cdot (x_i - x_c)}_{\textit{gradient level}} + (1-\theta)\cdot \underbrace{(\sum_{i=0}^{p_{S}-1}w_i\cdot x_i + w_c\cdot x_c)}_{\textit{intensity level}},
    \label{eq:gc-cdc}
\end{equation}
where $p_{S}$ is the number sampling pixels collected according to a certain direction (\ie, horizontal/vertical or diagonal) in the local region centered at $x_c$.

An additional benefit of C-CDC is either type of C-CDC has less parameters than CDC or standard convolution operator, as only half of the neighboring pixels are sampled.

\begin{figure}[t!]
\centering
    \centering
    \includegraphics[width=\linewidth]{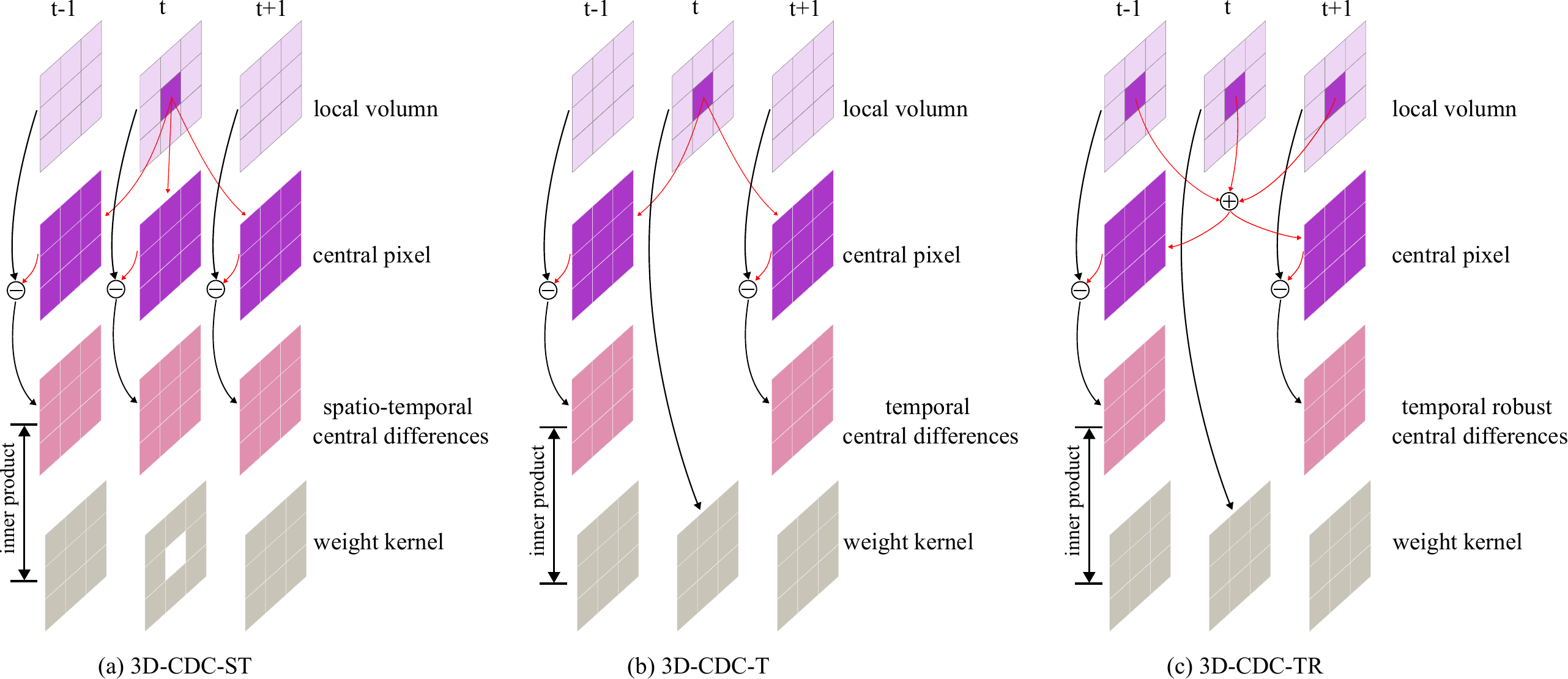}
    \caption{The design spirit of CDC can also be applied to 3D input, where the differences between pixels in other frames and pixels in current frame can be aggregated during 3D convolution to enrich temporal feature representation. The figure shows three versions of 3D-CDC. $\bigoplus$ denotes mean operation.}
    \label{fig:3d-cdc}
\end{figure}

\subsubsection{3D-CDC for spatial/temporal representation}
For video based visual tasks, spatio-temporal feature representation learning is the core to design well-performed CNN models. To exploit the rich local motion and enhance the spatio-temporal representation for 3D CNNs, 3D-CDC~\cite{yu2021-3d-cdc} was proposed to aggregate central differences along spatial and temporal directions as shown in~\cref{fig:3d-cdc}.

Via different subtraction strategies, 3D-CDC can be flexibly designed to capture spatial gradient information or temporal differences. Specifically, the authors in~\cite{yu2021-3d-cdc} developed three 3D-CDC operators for gesture recognition. First, spatio-temporal 3D-CDC (3D-CDC-ST), as shown in~\cref{fig:3d-cdc} (a), was designed considering both spatial and temporal gradient cues to effectively enhance local texture and motion details in RGB sequences. The second operator illustrated in~\cref{fig:3d-cdc} (b), temporal 3D-CDC (3D-CDC-T) was then proposed to focus on temporal central differences to better model temporal dynamics. Finally, considering sensor noise especially in the depth modality of RGB-D input, a more robust version of 3D-CDC (3D-CDC-TR) aims to reduce the sensitiveness to noise in pixels such as pixel jitters from the adjacent time steps by averaging the spatial centers of all time steps before calculating central differences (\cref{fig:3d-cdc} (c)).

The formulations of 3D-CDC can be easily derived by extending~\cref{eq:gcdc} or~\cref{eq:gc-cdc} to a 3D version which also uses the tradeoff parameter $\theta$ to combine both intensity and gradient level image cues.

\subsubsection{Discussion}
Unlike LBC~\cite{juefei2017lbc} that leverages binary sparse kernels to reduce computation and memory for convolutional operations (\ie, property \emph{\Rn{1}.} of LBP as we discussed in the beginning of this section), CDC series focus on property \emph{\Rn{2}.}: explicit encoding of higher-order information. In other words, rich gradient cues in both 2D images or 3D image sequences can be effectively captured by CDC to enhance visual feature representation, which are proved to be helpful for better prediction accuracy on various computer vision tasks.

\begin{figure}[t!]
\centering
    \centering
    \includegraphics[width=\linewidth]{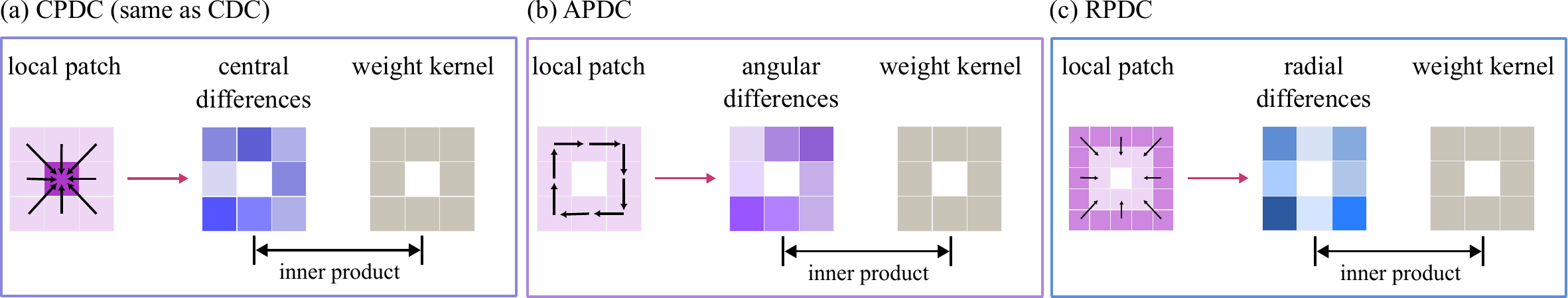}
    \caption{PDC is more general than CDC to probe local difference patterns in various encoding directions. The figure shows three versions of PDC, where CDC is a special case (a). By changing the sampling strategy, more PDC instances can be derived.}
    \label{fig:pdc}
\end{figure}

\subsection{Pixel difference convolution and its variants}

\subsubsection{Pixel difference convolution (PDC)}
PDC~\cite{su2021pdc} has a more general form in encoding local differences by changing the sampling strategies in the local region, allowing it to probe micro structures in a more flexible way, which is an essential property of LBP descriptors. 
In this way, the above introduced CDC becomes a special case of PDC where the central differences are adopted.

As shown in~\cref{fig:pdc} (a-c), PDC incorporates the calculation of pixel differences in a more general way when conducting convolutional operation. We can compare the standard convolution and PDC with the following formulations:

\begin{align}
    y_c &= f(\pmb{x}, \pmb{\theta}) = \sum_{i=1}^{p-1}w_{i}\cdot x_{i} + w_c\cdot x_c, \;\;\;\;\;\;\; \text{(standard convolution)} \label{eq:conv} \\
    y_c &= f(\Delta\pmb{x}, \pmb{\theta}) = \sum_{(x_i, x_i')\in \pmb{\mathcal{P}}}w_{i}\cdot (x_i - x_i'), \;\;\;\;\;\;\, \text{(PDC)} \label{eq:pdc}
\end{align}
where $\pmb{\mathcal{P}} = \{(x_1, x_1'), (x_2, x_2'), ..., (x_m, x_m')\}$ is the set of pixel pairs picked from the current local region and $m$ is the number of pixel pairs, $\pmb{\theta} = \{w_1, w_2, ..., w_i, ...\}$ are the kernel weights. 

To better capture diverse micro-structural patterns, pixel pairs can be selected according to various probing strategies in the LBP literature. In~\cite{su2021pdc}, LBP and ELBP \cite{ojala2002lbp2002riu2,liu2012extended,su2019bird} were adopted to encode pixel relations from varying directions (angular and radial). By integrating LBP and ELBP into convolution, three types of PDC instances were derived, namely, central PDC (CPDC), angular PDC (APDC), and radial PDC (RPDC), respectively (\cref{fig:pdc} (a-c)). For example, for APDC with a $3\times 3$ kernel, 8 pairs are first selected in the angular direction within the $3\times 3$ local region (thus $m=8$). Then,  pixel differences between these pixel pairs are aggregated with the kernel weights to calculate the output value. Obviously, both CDC and C-CDC are special cases of PDC by composing pixel pairs including the central pixel. In CDC, $m=8$ and in C-CDC, $m=4$. For both cases, $x_i'\equiv x_c$.

\begin{figure}[t!]
\centering
    \centering
    \includegraphics[width=\linewidth]{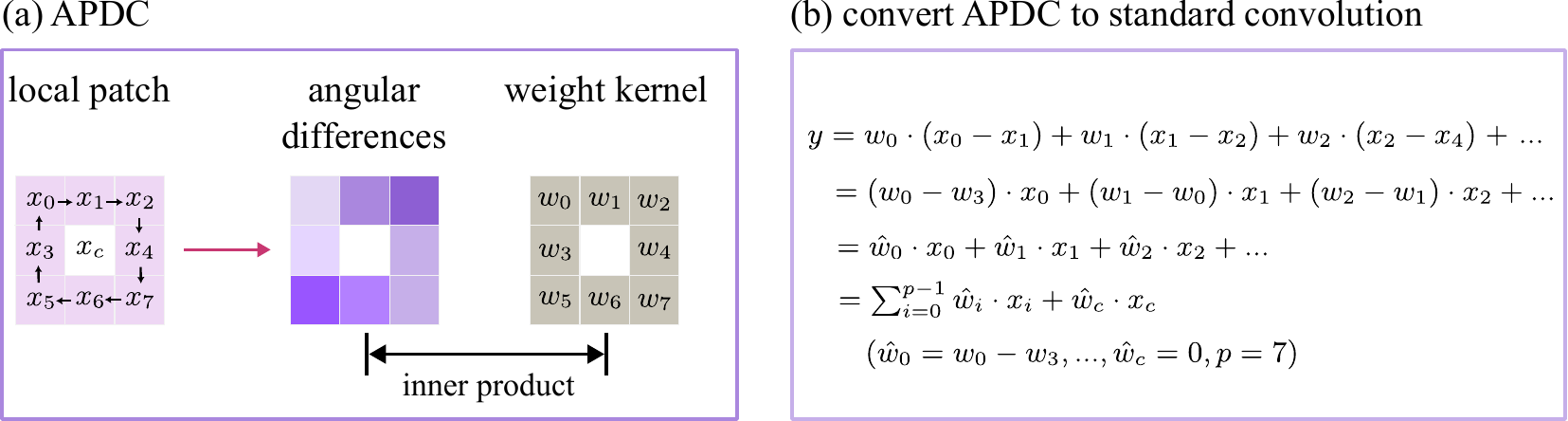}
    \caption{PDC can be converted to standard convolution to save computation and memory.}
    \label{fig:apdc}
\end{figure}

The authors in~\cite{su2021pdc} also proved that PDC can be converted to standard convolution with a novel reparameterization strategy to avoid double computation and runtime memory due to the calculation of pixel difference (\ie, \cref{eq:pdc} costs more computation and memory than~\cref{eq:conv}). An example of such conversion of APDC is illustrated in~\cref{fig:apdc}. By doing so, PDC is implemented as efficient as standard convolution during inference.




\subsubsection{Median pixel difference convolution (MeDiConv)}
Inspired by median robust extended LBP (MRELBP)~\cite{liu2016mrelbp} which improves LBP with greatly enhanced robustness to noise by considering local median values, MeDiConv~\cite{zhang2022mediconv} adopts a similar strategy when designing the convolution process. Specifically, by replacing $x_i'$ in~\cref{eq:pdc} with the local median value, MeDiConv can be written as:

\begin{equation}
    y_c = f(\Delta\pmb{x}_m, \pmb{\theta}) = \sum_{i=1}^{p-1}w_{i}\cdot (x_i - x_m), \;\;\;\;\;\;\, \text{(MeDiConv)} \label{eq: mediconv},
\end{equation}
where $x_m$ is the median value of the sampled local region.

Compared with standard convolution, MeDiConv is a nonlinear smoothing operation,
which can effectively remove outliers with limited impact on the ability of feature extraction. For standard convolution, noise corrupted image patterns could lead to false activations, leading to a significant decrease in accuracy. In contrast, for MeDiConv, the effect of applying MeDiConv at multiple layers can be considered as applying multiple median filters of different kernel sizes on the original image as each MeDiConv layer has a different receptive field. Consequently, MeDiNet was demonstrated to be able to deal with the noise of different levels.

\subsubsection{Discussion}
PDC inherits the properties \emph{\Rn{2}.} and \emph{\Rn{3}.} from LBP, forming a more versatile convolution operator to both capture gradient information from the image, and probing microstructural cues from different encoding directions. PDC enables a more straightforward way to integrate traditional LBP variant into modern CNN modules. Since both LBP and PDC involve sampling local pixels and selecting certain pixel differences, an LBP variant can be easily embedded into PDC when PDC adopts the same sampling and selection strategies following the corresponding LBP variant. MeDiConv is an example of such transfer from MRELBP to convolution, \ie, both calculate the differences between neighboring pixels and the local median value. Therefore, the noise robustness property of MRELBP is inherited in MeDiConv. Meanwhile, MeDiConv owns the feature extraction ability of CNNs. Benefits from both worlds are well combined.


\section{Applications}
\label{sec:app}

An extensive review on applications of traditional LBP has been already included in~\cite{liu2017lbpsurvey}. Therefore, we focus on providing a complementary review on applications of those LBP inspired CNN modules in the deep learning era.




{\small
\begin{longtable}{p{0.18\linewidth}|p{0.1\linewidth}|p{0.12\linewidth}|p{0.1\linewidth}|p{0.45\linewidth}}
\caption{Applications of LBP inspired CNN modules.}\label{tab:applications}\\
\toprule
Architecture name & Module & Application & Year & \multicolumn{1}{c}{Description} \\
\endfirsthead
\multicolumn{5}{c}%
{\tablename\ \thetable\ -- \textit{Applications of LBP inspired CNN modules (Continued from the previous page)}} \\
\toprule
Architecture name & Module & Application & Year & \multicolumn{1}{c}{Description} \\
\hline
\endhead
\hline \multicolumn{5}{r}{\textit{Continued on next page}} \\
\endfoot
\hline
\endlastfoot
\midrule
Local binary convolutional neural networks (LBCNN) \cite{juefei2017lbc} & LBC & Image classifica-tion & 2017 & By using LBC in AlexNet, LBCNN saves $6.622\times$ learnable parameters in the convolutional layers, while performing comparably to the original AlexNet on ImageNet dataset~\cite{deng2009imagenet}.\\
\midrule
Central Difference Convolutional Network (CDCN, CDCN++) \cite{yu2020cdc} & CDC & Face anti-spoofing & 2020 & CDCN is built by manually stacking multiple CDC layers; CDCN++ is searched using neural architecture search (NAS) technique where CDC operators with different channels become basic elements in the search space. It is demonstrated that standard convolution fails to capture the consistent spoofing pattern while CDC is able to extract the invariant detailed spoofing features, \eg, lattice artifacts. Comprehensive experiments are performed on six benchmark datasets to show that 1) CDC not only achieves superior performance on intra-dataset testing, 2) it also generalizes well on cross-dataset testing.\\
\midrule
Multi-Rate and Multi-Modal Temporal Enhanced Networks \cite{yu2021-3d-cdc} & 3D-CDC & Gesture recognition & 2021 & It is the first NAS based method for RGB-D gesture recognition. Similar to CDCN++, a novel search space is created by leveraging 3D-CDC-T and 3D-CDC-TR in the basic operators. 3D-CDC shows great ability to enhance the spatio-temporal representation for video understanding tasks. The resulting network achieves state-of-the-art performance on three benchmark datasets.\\
\midrule
Dual-Cross Central Difference Network (DC-CDN) \cite{yu2021c-cdc} & C-CDC & Face anti-spoofing & 2021 & Based on C-CDC operators, DC-CDN is established with cross feature
interaction modules for mutual relation mining and local detailed representation enhancement. Comprehensive experiments are performed on four benchmark datasets with three testing protocols to demonstrate the state-of-the-art performance. \\
\midrule
Pixel Difference Networks (PiDiNet) \cite{su2021pdc} & PDC & Edge detection & 2021 & PiDiNet is structured by sequentially stacking different PDC instances as well as the standard convolution to capture gradient information in different encoding directions. Since PDC can be converted to standard convolution, PiDiNet does not suffer from extra computation and memory consumption. With an efficient design of backbone and side structures, PiDiNet is the first deep network that can achieve human-level performance without ImageNet pretraining, when evaluated on popular edge detection datasets.\\
\midrule
Median Pixel Difference Convolutional Network (MeDiNet)~\cite{zhang2022mediconv} & MeDi-Conv & Face recognition & 2021 & MeDiNet is built by embedding MeDiConv into CNN architectures to equip CNNs with built-in robustness to noise of different levels. MeDiNet is tested on popular face datasets with challenging settings by adding different levels of noise (\eg, by changing blur kernels, noise intensities, scales, and JPEG quality factors). Extensive experiments show that MeDiNet can effectively remove noisy pixels in feature maps and suppress the negative impact of noise, leading to a more robust performance than standard CNNs.\\
\midrule
S-RaPiDiNet based on Random Pixel Different Convolution~\cite{liu2021random} & PDC & Face perception & 2021 & Unlike PDC adopting predefined sampling strategies for pixel pairs, Random PDC randomly samples pixel in the local region. The design mechanism is inspired by BRIEF descriptor~\cite{calonder2011brief}. Despite the simple strategy, S-RaPiDiNet shows great performance in face perception tasks.\\
\midrule
3D Central Difference Convolution Attention Network~\cite{zhao2021rppg} & 3D-CDC & rPPG measure-ment & 2021 & A central difference convolutional attention network for rPPG measurement is proposed. The adopted 3D-CDC can capture rich temporal context by gathering time difference information. \\
\midrule
Graph network based on Central Difference Graph Convolution (CDGC)~\cite{miao2021cdgc} & CDC & Action recognition & 2021 &  CDGC is designed for skeleton based action recognition. Like CDC aggregating the difference between neighboring pixels and the central pixel, CDGC takes differences between the features of the adjacent nodes and the central node. By generalizing CDGC with the incorporation of vanilla graph convolution, it is able to aggregate both node information and gradient information, leading to better accuracy than state-of-the-art methods. \\
\midrule
Multi-scale Texture Difference model (MTD-Net)~\cite{yang2021mtdnet} & CDC & Face forgery
detection & 2021 & MTD-Net considers both pixel intensity and pixel gradient information to give a stationary description of texture difference information by using CDC in the network.\\
\midrule
Gradient Siamese Network (GSN)~\cite{cong2022imagequality} & CDC & Image quality assessment & 2022 & Using CDC to obtain both semantic features and detail
difference hidden in image pair. GSN won the second place in NTIRE 2022 Perceptual Image Quality Assessment Challenge track 1 Full-Reference~\cite{gu2022ntire}.\\
\midrule
Suppression-Strengthen Network (S2N)~\cite{wan2022s2n} & CDC & Event-based recognition & 2022 & An evolution guided density-adaptive central difference convolution scheme is proposed to progressively encode the local center-surrounding variation and adaptively aggregate the features into a complete event representation under the guidance of the motion evolution map. \\ 
\midrule
Depth Dynamic Center Difference Convolution (DDCDC) based network~\cite{wu2022ddcdc} & CDC & Monocular 3D Object Detection & 2022 & DDCDC introduces surrounding pixel cues in depth estimation and has different convolution kernels weights for each pixel of all examples. DDCDC not only overcomes the limitations of conventional 2D convolution, but also highlights the differences in depth information between the target and the background, so more attention is paid to interesting objects\\
\midrule
Fast Saliency Model (FSM)~\cite{zabihi2022realsaliency} & CDC & Saliency prediction & 2022 & FSM consists of a modified U-net architecture, a location-dependent fully-connected layer, and CDC layers. Using the CDC layers at different scales enables capturing more robust and biologically motivated features. \\
\midrule
Local Motion and Contrast Priors Driven Deep Network (MoCoPnet)~\cite{ying2022mocopnet} & CDC & Infrared small target super-resolution & 2022 & Motivated by the local contrast prior in the spatial dimension, a central difference residual group is proposed to incorporate CDC into the feature extraction backbone, which can achieve center-oriented gradient-aware feature extraction to improve the target contrast for the task.\\
\midrule
CDC-based Multi-Receptive-Field (CDC-MRF) module~\cite{CAO2022} & CDC & Image enhance-ment & 2022 & CDC is embedded in the architecture to effectively extract multi-scale edge/texture features on thermal images. The extracted thermal features are then utilized as important guidance to facilitate the following low-light visible image enhancement task.\\
\midrule
RPDC based network structure~\cite{xu2022yarn} & PDC & yarn contour detection & 2022 & Firstly, according to the directional gradient distribution of yarn images in the visual system, a radial pixel difference convolution is improved to extract interested edge features. Secondly, considering the graphical characteristics of the yarn to be inspected, a mask layer of texture erosion is designed to further filter irrelevant details from extracted edge features.\\
\midrule
Semantic Diffusion Network (SDN)~\cite{tan2022sdn} & CDC & Semantic segmenta-tion & 2022 & SDN is proposed for approximating the diffusion process, which contains a parameterized semantic difference convolution operator followed by a feature fusion module and constructs a differentiable mapping from original backbone features to advanced boundary-aware features. \\
\bottomrule
\end{longtable}
}

\cref{tab:applications} lists the methods published in recent years that leverage LBP inspired CNN modules on a wide range of applications, including image classification~\cite{juefei2017lbc}, face anti-spoofing~\cite{yu2020cdc,yu2021c-cdc}, gesture recognition~\cite{yu2021-3d-cdc}, edge detection~\cite{su2021pdc}, facial analysis~\cite{zhang2022mediconv,liu2021random,yang2021mtdnet}, remote photoplethysmography (rPPG) measurement~\cite{zhao2021rppg}, action recognition~\cite{miao2021cdgc}, image quality accessment~\cite{cong2022imagequality}, event-based recognition~\cite{wan2022s2n}, monocular object detection~\cite{wu2022ddcdc}, saliency detection~\cite{zabihi2022realsaliency}, infrared target super-resolution~\cite{ying2022mocopnet}, and semantic segmentation~\cite{tan2022sdn}. Here, it would be tricky to compare different methods with quantitative evaluations due to the big variation of tasks. Generally, the three properties of LBP have been well-integrated into CNN architectures and shown significant advantages in boosting convolutional architectures for various of computer vision tasks.

\section{Future works and conclusion}
\label{sec:conclusion}
\subsubsection{Future works}
Although great success has been made during the last years on developing novel modules with LBP, which enhance CNNs in efficiency, representational capacity, and discriminative power, we believe that it is just a starting step and the following aspects are worth exploring in the future.
\begin{itemize}
    \item Currently, most LBP inspired deep modules are based on CNN architectures (\eg, CDC, PDC) to extract gradient information on regular data like images and videos. However, gradient information is also important for irregular data like point clouds \cite{Guo2021Pointcloudssurvey,su2022svnet} and 3D meshes~\cite{chen20213dmeshs}. Designing novel modules that benefit irregular data can be a valuable direction.
    \item The three properties we discussed in~\cref{sec:cnn-lbp} are the main motivations when designing LBP inspired modules. However, none of the existing methods utilizes all the three properties in tandem, limiting the current modules to fully enjoy the benefits of LBP. Further innovation can be made to meet such inspiring target. For example, fusing PDC to binary neural networks~\cite{courbariaux2016bnn,su2020ftbnn,zhang2022dynamic}. 
    \item PDC enjoys the general form to organize the encoding of pixel differences, making it possible to incorporate any LBP variant to the convolutional module. While only the orignial LBP~\cite{ojala2002lbp2002riu2} and ELBP variants~\cite{liu2012extended} are now explored and show positive effect on the task of edge or contour detection~\cite{su2021pdc,xu2022yarn}, we believe more LBP variants can be considered to tackle other different vision tasks. 
    \item Like 3D-CDC, PDC can also be generalized to 3D scenes, where the temporal difference cues can be extracted in a more flexible way, rather than only considering the central differences.
\end{itemize}

\subsubsection{Conclusion}
This paper presents a review on LBP and the derived CNN modules proposed in the literature. We provide a comprehensive study on how traditional LBP and its variants inspired the design of modern CNN modules to boost the performance of vision models in the deep learning era. The associated applications in recent years are also elaborated. Finally, we list possible future directions. As an example of combining traditional vision  descriptors with deep learning, we believe our review can also encourage more researchers to rethink the role of traditional descriptors when developing deep learning methods.

\section*{Acknowledgement}
This work was partially supported by National Key Research and Development Program\\of China No. 2021YFB3100800, the Academy of Finland under grant 331883, Infotech Project FRAGES,
and the National Natural Science Foundation of China under Grant 62022091 and 62201588.

%
%
%
\bibliographystyle{splncs04}
\bibliography{paper_44}
\end{document}